\documentclass{article}

    \usepackage[preprint]{neurips_2024}

\usepackage[utf8]{inputenc} % allow utf-8 input
\usepackage[T1]{fontenc}    % use 8-bit T1 fonts
\usepackage{hyperref}       % hyperlinks
\usepackage{url}            % simple URL typesetting
\usepackage{booktabs}       % professional-quality tables
\usepackage{amsfonts}       % blackboard math symbols
\usepackage{nicefrac}       % compact symbols for 1/2, etc.
\usepackage{microtype}      % microtypography
\usepackage{xcolor}         % colors

\usepackage{enumitem}
\usepackage{amsmath}
\usepackage{multirow}
\usepackage{adjustbox}
\usepackage{array}
\usepackage{booktabs} % for professional tables
\usepackage{tabularx}
\usepackage{graphicx}

\usepackage{natbib}
\setcitestyle{numbers}

\usepackage{inconsolata}
\newcommand\blfootnote[1]{%
\begingroup
\renewcommand\thefootnote{}\footnote{#1}%
\addtocounter{footnote}{-1}%
\endgroup
}

\title{Step-Controlled DPO: Leveraging Stepwise Error for Enhanced Mathematical Reasoning}

\author{%
  Zimu Lu, Aojun Zhou, Ke Wang, Houxing Ren, Weikang Shi\\ {\bf Junting Pan}, {\bf Mingjie Zhan}$^{\dagger}$, {\bf Hongsheng Li}$^{\dagger}$\\
  Multimedia Laboratory (MMLab), The Chinese University of Hong Kong\\
 \texttt{luzimu@mail.ustc.edu.cn} \quad \texttt{\{aojunzhou, zmjdll\}@gmail.com}  \\ \texttt{hsli@ee.cuhk.edu.hk} 
}

\begin{document}

\maketitle

\begin{abstract}

Direct Preference Optimization (DPO) has proven effective at improving the performance of large language models (LLMs) on downstream tasks such as reasoning and alignment. In this work, we propose Step-Controlled DPO (SCDPO), a method for automatically providing stepwise error supervision by creating negative samples of mathematical reasoning rationales that start making errors at a specified step. By applying these samples in DPO training, SCDPO can better align the model to understand reasoning errors and output accurate reasoning steps. We apply SCDPO to both code-integrated and chain-of-thought solutions, empirically showing that it consistently improves the performance compared to naive DPO on three different SFT models, including one existing SFT model and two models we finetuned. Qualitative analysis of the credit assignment of SCDPO and DPO demonstrates the effectiveness of SCDPO at identifying errors in mathematical solutions. We then apply SCDPO to an InternLM2-20B model, resulting in a 20B model that achieves high scores of 88.5\% on GSM8K and 58.1\% on MATH, rivaling all other open-source LLMs, showing the great potential of our method.  Related code for data generation and training is released at \url{https://github.com/mathllm/Step-Controlled_DPO}.

\end{abstract}
\blfootnote{$^\dagger$Corresponding author}

\section{Introduction}

% introduce DPO for mathematical reasoning
Large language models (LLMs) have shown great potential in mathematical problem-solving. Recently, Direct Preference Optimization (DPO;~\cite{rafailov2024direct}) has emerged as a popular choice for aligning LLMs with relative feedback to improve the quality of generated text. Prior works~\cite{christiano2017deep, pal2024smaug, xu2024chatglm} have demonstrated that reinforcement learning algorithms and DPO can improve the mathematical reasoning abilities of LLMs, making the generated reasoning process more controllable. Different from other tasks that need human or AI feedback, the final answer to a mathematical problem serves as a reliable way to judge the quality of the model's response, since a mathematical problem typically has a single correct answer. As a result, the responses producing the correct final answers are desirable and can serve as the preferred samples, while the ones reaching incorrect final answers are undesirable and can serve as the dispreferred samples.

% introduce SCDPO
However, solutions to a mathematical problem can be diverse, with many different reasoning paths arriving at the correct final answer and many subtle ways to make mistakes. Determining the preferred and dispreferred responses based on the final answer is coarse and may be inadequate for capturing \textit{the intricacies of the multi-step mathematical reasoning process.} Previous studies introduce process supervision~\cite{lightman2023let}, but it requires large amounts of meticulous and expensive human annotation and only applies to traditional RL algorithms.

In this paper, we show how to automatically provide explicit stepwise preference supervision by generating dispreferred solutions that start making errors at a specific step. We propose \textit{Step-Controlled DPO (SCDPO)}, an algorithm that introduces stepwise supervision without necessitating extra human annotation. This approach starts with a model finetuned with question-solution pairs and possessing initial math-solving capabilities, which is used to generate solutions to a set of math problems. We choose the solutions whose final answers match those of the ground truth. The reasoning steps in these solutions can be seen as correct, since the cases of a wrong solution reaching the right answer are rare. We take each of these correct solutions and start generating with the model via modulating the hyperparameter of the model, i.e., increasing the temperature of the final softmax function, from various intermediate steps of that solution, and retain the samples where the final answer is incorrect. In this way, the steps before the intermediate step are the same as the original correct solution, while the steps after are the ones with possible errors. During DPO training, the correct solutions are the preferred samples, and they are paired with the wrong solutions generated in this way, with the question and the steps before the intermediate step as the prompts. These step-controlled training samples help models learn detailed reasoning abilities and are mixed with naive DPO training data produced by only checking the final answer, which optimizes the general form of the solution. 

% state contributions
Our contributions are as follows:

\begin{itemize}[leftmargin=*]
  \item We introduce Step-Controlled DPO (SCDPO), which we empirically show improves the performance of DPO in enhancing LLMs' mathematical reasoning abilities. We also conduct qualitative analysis of credit assignment of SCDPO.
  \item We conduct experiments on chain-of-thought and code-integrated solutions, showing that SCDPO can effectively improve mathematical problem-solving performance of three different SFT models.
  \item Using SCDPO, we finetune an InternLM2-20B model, which reaches 88.5\% on GSM8K~\cite{cobbe2021training} and 58.1\% on MATH~\cite{hendrycks2021measuring}, rivaling all other open-source models, demonstrating the great potential of our method.
\end{itemize}

\section{Related Work}

\textbf{LLM for Mathematical Reasoning.} Prior works have explored various methods to enhance mathematical reasoning abilities of LLMs. Prompting methods, such as Chain-of-Thought~\cite{wei2022chain}, Tree-of-Thought~\cite{yao2024tree}, PAL~\cite{gao2023pal}, Program-of-Thought~\cite{chen2022program}, and CSV~\cite{zhou2023solving}, use carefully engineered prompts to bring out LLMs' mathematical skills without changing their parameters. Other works optimize parameters of LLMs for enhanced mathematical reasoning through either pretraining or finetuning. Llemma~\cite{azerbayev2023llemma}, and MathPile~\cite{wang2023generative} continue pretraining LLMs on large amounts of math-related data, while RFT~\cite{yuan2023scaling}, Mammoth~\cite{yue2023mammoth}, MathCoder~\cite{wang2023mathcoder}, WizardMath~\cite{luo2023wizardmath}, ToRA~\cite{gou2023tora}, MetaMath~\cite{yu2023metamath}, MathGLM~\cite{yang2023gpt}, and MathGenie~\cite{lu2024mathgenie} finetune pretrained models on question-solution pairs. These methods effectively improves LLMs' ability to solve challenging mathematical problems, demonstrating impressive performance on mathematical benchmarks such as GSM8K~\cite{cobbe2021training}, MATH~\cite{hendrycks2021measuring}, etc. Our work builds upon models that have undergone pretraining and finetuning, using DPO to further enhance their mathematical abilities.

\textbf{Aligning LLMs Using Relative Feedback.} Methods that align LLMs with human or AI annotated preference data have been used to improve performance on a variety of downstream tasks such as translation~\cite{kreutzer2018reliability}, summarization~\cite{stiennon2020learning,ziegler2019fine}, and instruction-following~\cite{ouyang2022training,ramamurthy2022reinforcement}. Reinforcement learning from human (or AI) feedback~\cite{christiano2017deep, bai2022constitutional} first trains a reward model, then uses reinforcement learning algorithms such as REINFORCE~\cite{Williams2004SimpleSG}, PPO~\cite{schulman2017proximal}, or variants~\cite{ramamurthy2022reinforcement} to maximize the reward. To simplify the pipeline, several direct alignment methods~\cite{rafailov2024direct,azar2024general,zhao2023slic} have been proposed. Among them, DPO~\cite{rafailov2024direct} and several of its variants~\cite{pal2024smaug,ethayarajh2024kto,liu2023statistical,azar2024general} offer a way to optimize the reward function without having to train an extra reward model, proving highly effective on various tasks~\cite{tunstall2023zephyr,yuan2024self}. 

Recently, these preference alignment methods have also been applied to mathematical problem-solving tasks. DeepSeekMath~\cite{shao2024deepseekmath} and R$^3$~\cite{xi2024training} uses RL to improve mathematical accuracy, while ChatGLM-Math~\cite{xu2024chatglm} and Process Reward Synthesizing~\cite{jiao2024learning} uses DPO to improve model's mathematical generation quality. Process supervision~\cite{lightman2023let,wang2023math} uses stepwise preference of mathematical solution in its RL finetuning, which is highly effective but needs costly fine-grained human annotation. Our work offers a way to create stepwise error annotations of preferred and dispreferred solution pairs, and uses the data to improve DPO's performance on mathematical problem-solving tasks.

\section{Step-Controlled DPO Pipeline}

In this section, we introduce Step-Controlled DPO (SCDPO), a pipeline for automatically generating preferred and dispreferred responses to math problems, with annotations of erroneous solving steps, and using these responses in DPO training to enhance the mathematical reasoning abilities of LLMs. Our method consists of two stages: step-controlled data generation, and step-aware DPO training. The two stages construct a feedback-alignment framework that is both effective and cost-efficient.

\textbf{Initial Model.} Our method starts with an initial model, denoted as $\pi_{\text{SFT}}$, which has been finetuned with question-solution pairs from the training sets of GSM8K and MATH, two high-quality mathematical datasets that contain grad-school math word problems and competition-level math problems, respectively. When prompted with a math problem $q$, $\pi_{\text{SFT}}$ is able to generate a step-by-step solution, denoted as $a$. $a$ can be broken down into a sequence of reasoning steps, for example, $a=(t_0, \ldots, t_m)$. Here, $t_i$ $(i = 0, \ldots, m)$ represents either a code reasoning step or a natural language reasoning step within $a$.

We experiment with two solution formats: code-integrated solution format~\cite{zhou2023solving}, and chain-of-thought solution format~\cite{wei2022chain}. For the code-integrated solution format, $t_i$ in $(t_0, \ldots, t_m)$ alternates between a code reasoning step and a natural language reasoning step, while for the chain-of-thought solution format, all the steps are in natural language. We primarily use the code-integrated solution format, as previous works~\cite{wang2023mathcoder,lu2024mathgenie,gou2023tora} show that it results in higher accuracy than chain-of-thought. We finetune a Mistral-7B model with 34K samples of code-integrated solutions from GSM8K and 47K from MATH to create the initial model. We also validate our method on the chain-of-thought format using the off-the-shell MetaMath-Mistral-7B model, as well as MathCoder-Mistral-7B, which we trained using the MathCodeInstruct dataset~\cite{wang2023mathcoder}.

\subsection{Step-Controlled Data Generation}
\label{sec:scdpo_data_collection}

\begin{figure*}[t]
    \centering
    \includegraphics[width=1.0\textwidth]{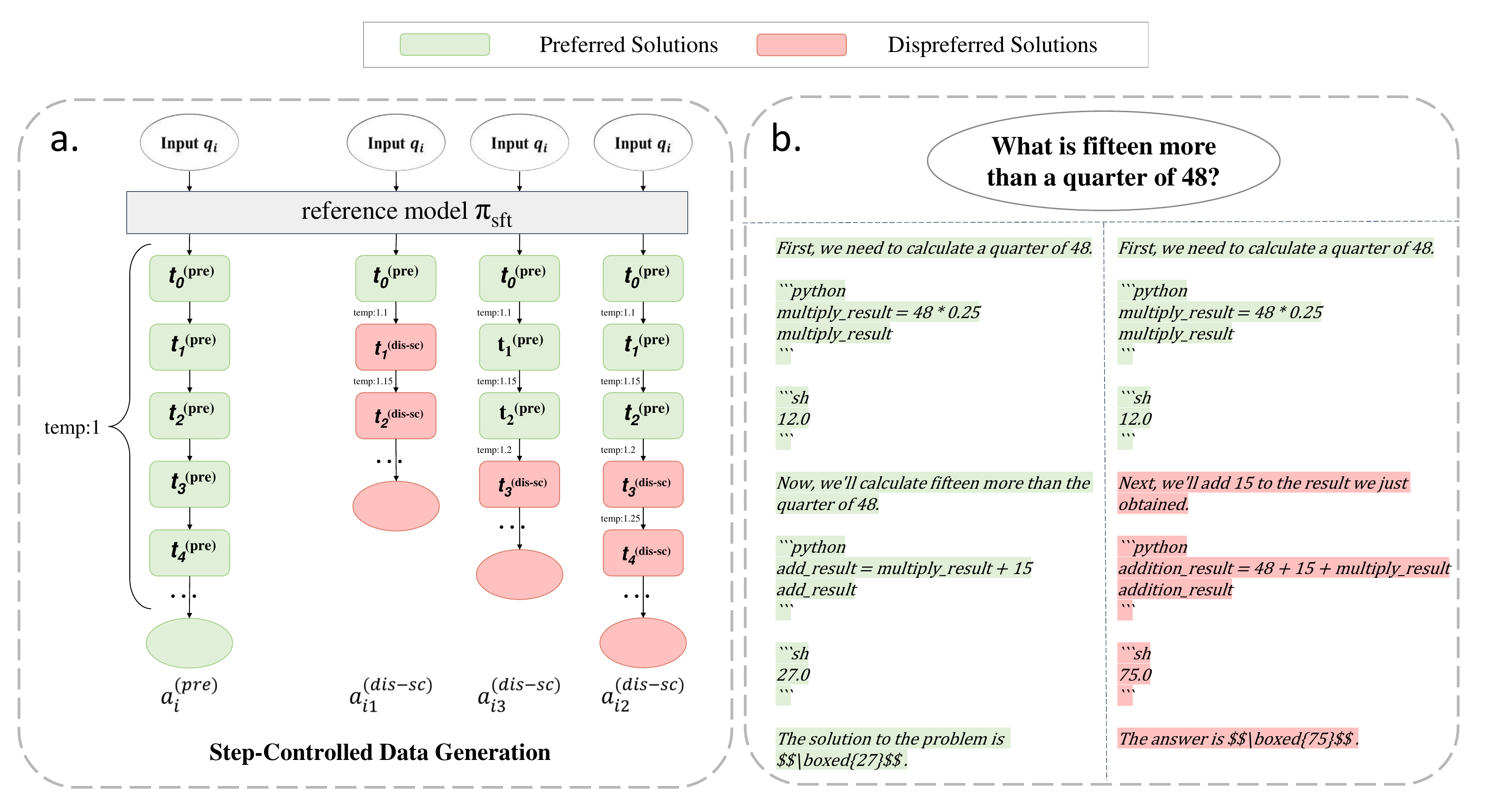}
    \caption{Demonstration and example of the step-controlled data generation process. \textbf{a.} Step-controlled data generation. First, a solution reaching the correct final answers is collected, which we denote as $a_i^\text{(pre)}$. Then, erroneous solutions that reach incorrect final answers are generated, starting from intermediate steps of $a_i^\text{(pre)}$, creating dispreferred solutions $a_{i1}^\text{(dis-sc)}$, $a_{i2}^\text{(dis-sc)}$, and $a_{i3}^\text{(dis-sc)}$. These dispreferred solutions share the steps before the intermediate steps with $a_i^\text{(pre)}$. The temperature of the newly generated steps gradually increases with each step to make the generation more erroneous. \textbf{b.} An example of a pair of preferred and dispreferred solutions. The dispreferred solution starts making errors after a particular intermediate step.}
    
\label{fig:step_controlled_data_collection}
\end{figure*}

The data we collect is in two parts: naive DPO data $D_\text{naive}$ and Step-Controlled DPO data $D_\text{SC}$.

\textbf{Generation of $D_\text{naive}$.} $D_\text{naive}$ contains pairs of preferred-dispreferred samples, used to optimize the general form of the solution. To create $D_\text{naive}$, we prompt $\pi_{\text{SFT}}$ with math questions in the training sets of GSM8K and MATH. The training set of GSM8K and MATH each contains 7.5K questions. Each question is presented to $\pi_{\text{SFT}}$ multiple times and various solutions are generated, with a temperature of 1. The quality of the generated solutions is judged by the final answers. If a solution reaches the same final answer as the ground truth, and no errors or adjustments occur at any of the reasoning steps (we detect these by looking for strings like ``error'' or ``apologies''), the solution is seen as preferred, while the solutions that reach answers different from the ground truth are considered dispreferred. The questions in GSM8K and MATH are all open-ended, and it is unlikely for incorrect reasoning steps without errors and adjustments to lead to the correct answer, so the reasoning steps in the preferred solutions can reliably be seen as correct. We randomly sampled 87 solutions that reach correct final answers, and found that of the 369 reasoning steps in these solutions, only 2 contain errors, which is a very small percentage (about 0.5\%). The solution generation of each question stops when at least one preferred solution and one dispreferred solution are generated, or the number of solutions generated reaches an upper limit of $K$. For the code-integrated solution format, $K$ is set as 100. This results in around 6.5K preferred-dispreferred solution pairs for GSM8K and MATH respectively, combining into a total of approximately 13K DPO training pairs. The resulting data can be expressed as:

\begin{align}
D_\text{naive} = \{(q_i, a_i^\text{(pre)}, a_i^\text{(dis)}): i=1, \ldots, N_\text{naive}\}
\end{align}

Here, $q_i$ denotes the $i$th question, while $a_i^\text{(pre)}$ and $a_i^\text{(dis)}$ represent the preferred and dispreferred solution to the $i$th question.

\textbf{Generation of $D_\text{SC}$.} In order to generate solutions with stepwise error annotations for DPO training, we propose a method to automatically generate training data with errors starting to occur at a controlled step. The process is demonstrated in Fig.~\ref{fig:step_controlled_data_collection}. To do this, we first take a preferred solution from $D_\text{naive}$, denoted as $a_i^\text{(pre)} = (t_0^\text{(pre)}, \ldots, t_k^\text{(pre)}, t_{k+1}^\text{(pre)}, \ldots, t_{m_i}^\text{(pre)})$. Here, $t_k^\text{(pre)}$ is a random intermediate step within $a_i^\text{(pre)}$. As $a_i^\text{(pre)}$ is a correct solution, $t_0^\text{(pre)}, \ldots, t_{m_i}^\text{(pre)}$ are all correct steps, as we have taken care to retain only those solutions with no execution errors, apologies or rectifications. As shown in Fig.~\ref{fig:step_controlled_data_collection} a, to create a solution with errors occurring after step $k$, we present $\pi_{\text{SFT}}$ with sequence $(q_i, t_0^\text{(pre)}, \ldots, t_k^\text{(pre)})$, and raise the temperature of the final softmax function to affect the generation quality, increasing the occurrence of errors in the following steps. Raising the temperature causes the model performance to become unstable and erroneous. The effect of raised temperature on accuracy is demonstrated in Fig.~\ref{fig:different_temp_acc} of Appendix.~\ref{sec:different_temp_acc}. We observe that when the temperature is instantly raised and remains at a high value, the model can generate garbled strings as errors accumulate, which does not represent any reasoning mistakes and contains no valuable information. To avoid this, we adopt a gradually increasing temperature, which initially starts at 1.1, and increases by 0.05 with each generated step, until the generation ends or the temperature reaches 1.4. This setting empirically reduces the frequency of the occurrence of garbled text, while increasing the error rate. We generate the steps following $(q_i, t_0^\text{(pre)}, \ldots, t_k^\text{(pre)})$ multiple times, until one reaching an incorrect answer is generated. Appending the generated steps to $(t_0^\text{(pre)}, \ldots, t_k^\text{(pre)})$, we get a dispreferred solution with step-controlled error, denoted as $a_{ik}^{\text{(dis-sc)}} = (t_0^\text{(pre)}, \ldots, t_k^\text{(pre)}, t_{k+1}^\text{(dis-sc)}, \ldots, t_{n_i}^\text{(dis-sc)})$, where the sequence $(t_{k+1}^\text{(dis-sc)}, \ldots, t_{n_i}^\text{(dis-sc)})$ is erroneous. An example is presented in Fig.~\ref{fig:step_controlled_data_collection} b. The resulting data can be expressed as:

\begin{align}
D_\text{SC} = \{(q_{i}, a_i^\text{(pre)}, a_{ik}^\text{(dis-sc)}): i=1, \ldots, N_\text{SC}; k\in[0, m_i - 1]\}
\end{align}

Here, $q_i$ denotes the $i$th question, while $a_i^\text{(pre)}$ is the preferred solution, and $a_{ik}^{\text{(dis-sc)}}$ is the dispreferred solution with step-controlled error that occurs after $t_k^\text{(pre)}$. $N_\text{SC}$ is the number of questions in $D_\text{SC}$, while $m_i$ is the index of the last step of $a_i^\text{(pre)}$. 

\subsection{Step-Controlled DPO Training}

Having collected $D_\text{naive}$ and $D_\text{SC}$, we apply them to DPO training. $D_\text{naive}$ serves to regulate the general form of solutions, while $D_\text{SC}$ supervises the model's reasoning on a step level. During DPO training, samples in $D_\text{naive}$ and $D_\text{SC}$ are mixed together randomly, and the DPO loss is applied to each sample. For samples from $D_\text{naive}$, the loss is applied to all steps in the preferred and dispreferred solutions, which can be written as:

\begin{align}
&\mathcal{L}_{\rm naive}(\pi_\theta; \pi_\text{SFT}) \nonumber\\
&= -\mathbb{E}_{(q_i, a_i^\text{(pre)}, a_i^\text{(dis)})\sim \mathcal{D_{\text{naive}}}} 
\left[ \log \sigma \left( \beta \log \frac{\pi_\theta(a_i^\text{(pre)} | q_i)}{\pi_\text{SFT}(a_i^\text{(pre)} | q_i)} - \beta \log \frac{\pi_\theta(a_i^\text{(dis)} | q_i)}{\pi_\text{SFT}(a_i^\text{(dis)} | q_i)} \right) \right]
\end{align}

For a pair of preferred and dispreferred solutions in $D_\text{SC}$,  $a_i^\text{(pre)} = (t_0^\text{(pre)}, \ldots, t_k^\text{(pre)}, t_{k+1}^\text{(pre)}, \ldots, t_{m_i}^\text{(pre)})$ and $a_{ik}^{\text{(dis-sc)}} = (t_0^\text{(pre)}, \ldots, t_k^\text{(pre)}, t_{k+1}^\text{(dis-sc)}, \ldots, t_{n_i}^\text{(dis-sc)})$, the loss is only applied to the steps after $t_k^\text{(pre)}$, i.e. $(t_{k+1}^\text{(pre)}, \ldots, t_{m_i}^\text{(pre)})$ and $(t_{k+1}^\text{(dis-sc)}, \ldots, t_{n_i}^\text{(dis-sc)})$. We denote $(t_0^\text{(pre)}, \ldots, t_k^\text{(pre)})$ as $a_{ik-\text{front}}^\text{(pre)}$, $(t_{k+1}^\text{(pre)}, \ldots, t_{m_i}^\text{(pre)})$ as $a_{ik-\text{end}}^\text{(pre)}$, and $(t_{k+1}^\text{(dis-sc)}, \ldots, t_{n_i}^\text{(dis-sc)})$ as $a_{ik-\text{end}}^\text{(dis-sc)}$, so $a_i^\text{(pre)} = (a_{ik-\text{front}}^\text{(pre)}, a_{ik-\text{end}}^\text{(pre)})$, and $a_{ik}^\text{(dis-sc)} = (a_{ik-\text{front}}^\text{(pre)}, a_{ik-\text{end}}^\text{(dis-sc)})$. The loss function can be written as:

\begin{align}
&\mathcal{L}_{\rm SC}(\pi_\theta; \pi_\text{SFT}) = \nonumber\\
&\resizebox{.94\hsize}{!}{$-\mathbb{E}_{(q_i, a_i^\text{(pre)}, a_{ik}^\text{(dis-sc)})\sim \mathcal{D_{\text{SC}}}} 
\left[ \log \sigma \left( \beta \log \frac{\pi_\theta(a_{ik-\text{end}}^\text{(pre)} | q_i, a_{ik-\text{front}}^\text{(pre)})}{\pi_\text{SFT}(a_{ik-\text{end}}^\text{(pre)} | q_i, a_{ik-\text{front}}^\text{(pre)})} - 
\beta \log \frac{\pi_\theta(a_{ik-\text{end}}^\text{(dis-sc)} | q_i, a_{ik-\text{front}}^\text{(pre)})}{\pi_\text{SFT}(a_{ik-\text{end}}^\text{(dis-sc)} | q_i, a_{ik-\text{front}}^\text{(pre)})} \right) \right]$}
\end{align}

Combining $\mathcal{L}_{\rm naive}$ and $\mathcal{L}_{\rm SC}$, the final loss function of Step-Controlled DPO is as follows:

\begin{align}
& \mathcal{L}_{\rm SCDPO} = \mathcal{L}_{\rm naive} + \mathcal{L}_{\rm SC}
\end{align}

In this way, $\mathcal{L}_{\rm naive}$ optimizes the general form of the solution, while $\mathcal{L}_{\rm SC}$ focuses on detailed reasoning steps, thus improving the model's accuracy in solving mathematical problems.

\section{Theoretical Explanation of Step-Controlled DPO}

\textbf{Theoretical Insight.}
In this section, we provide some theoretical insights into why SCDPO can effectively enhance the reasoning ability of LLMs. As explained in~\cite{rafailov2024r}, the DPO loss can be cast into token-level MDP. Similarly, we can also derive a step-level MDP for $\mathcal{L}_{\rm SC}$ as follows:

\begin{align}
&\mathcal{L}_{\rm SC}(\pi_\theta; \pi_\text{SFT}) =\nonumber\\
&\resizebox{.94\hsize}{!}{$-\mathbb{E}_{(q_i, a_i^\text{(pre)}, a_{ik}^\text{(dis-sc)})\sim \mathcal{D_{\text{SC}}}} 
\left[ \log \sigma \left( \left(\sum\limits_{j=k+1}^{m_i}\beta \log \frac{\pi_\theta(t_j^{\text{(pre)}} | q_i, t_{<j})}{\pi_\text{SFT}(t_j^{\text{(pre)}} | q_i, t_{<j})}\right) - 
\left(\sum\limits_{j=k+1}^{n_i}\beta \log \frac{\pi_\theta(t_j^{\text{(dis-sc)}} | q_i, t_{<j})}{\pi_\text{SFT}(t_j^{\text{(dis-sc)}} | q_i, t_{<j})}\right) \right) \right]$}
\end{align}

Here, $\beta \log \frac{\pi_\theta(t_j^{\text{(pre)}} | q_i, t_{<j})}{\pi_\text{SFT}(t_j^{\text{(pre)}} | q_i, t_{<j})}$ and $\beta \log \frac{\pi_\theta(t_j^{\text{(dis-sc)}} | q_i, t_{<j})}{\pi_\text{SFT}(t_j^{\text{(dis-sc)}} | q_i, t_{<j})}$ represent the reward of a single preferred or dispreferred step. For naive DPO, all steps in the preferred and dispreferred solutions have their rewards affecting the loss. However, many steps in the dispreferred solution are actually correct, as the error often occurs in a later step. Step-Controlled DPO reduces the range of steps, starting from the $(k+1)$th step, from which the dispreferred steps are more likely to be erroneous due to the raised sampling temperature. The focus of the optimization is thus cast on the errored steps rather than the whole solution, letting the model learn more detailed reasoning abilities.

\textbf{Qualitative Evaluation of Credit Assignment of SCDPO.}
We perform qualitative evaluation of credit assignment on two models trained with SCDPO and DPO respectively. For a sequence of tokens $\mathbf{x}=(x_0, \ldots, x_m)$, where $x_i$ is the $i$th token in the sequence, we denote all the tokens before $x_i$ as $\mathbf{s}_i$, written as $\mathbf{s}_i=(x_0, \ldots, x_{i-1})$. As introduced in recent research~\cite{yu2023metamath}, the DPO implicit reward can be expressed as follows:

\begin{equation}\label{eq:reward_param}
    r(\mathbf{s}_i, x_i) = \beta \log \pi(x_i| \mathbf{s}_i) - \beta \log \pi_\text{SFT}(x_i| \mathbf{s}_i)
\end{equation}

\begin{figure*}[t]
    \centering
    \includegraphics[width=\linewidth]{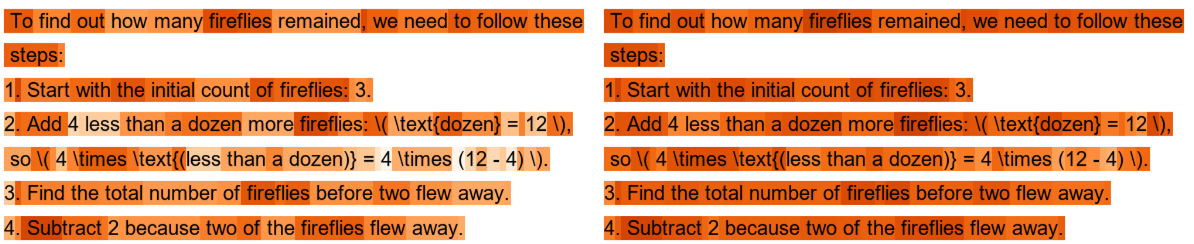}
    \caption{Credit assignment of part of a solution for a GSM8K problem. Each token is colored corresponding to the DPO implicit reward as expressed in Eq.~\ref{eq:reward_param} (darker is higher). The left is the credit assignment of SCDPO, which correctly highlights the error -- 4 less than a dozen is not 4 times (12 - 4), while the credit assignment of DPO on the  right fails to highlight it.}
    
\label{fig:rewards_gsm8k_1}
\end{figure*}

\begin{figure*}[t]
    \centering
    \includegraphics[width=\linewidth]{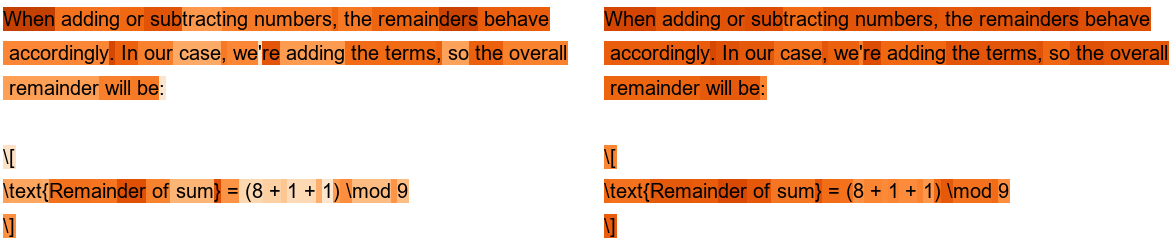}
    \caption{Credit assignment of part of a solution for a MATH problem. Each token is colored corresponding to the DPO implicit reward as expressed in Eq.~\ref{eq:reward_param} (darker is higher). The left is the credit assignment of SCDPO, which correctly highlights the error -- as the original question was ``Find the remainder when $8\cdot10^{18}+1^{18}$ is divided by 9'', the remainders of the terms $8$, $10^{18}$, and $1^{18}$ should not be summed, while the credit assignment of DPO on the  right fails to highlight the error.}
    
\label{fig:rewards_math_1}
\end{figure*}

Here $r(\mathbf{s}_i, x_i)$ denotes the DPO implicit reward of token $x_i$, which is the value we visualize as the background color of the token. A darker color represents a higher reward value. As demonstrated in Fig.~\ref{fig:rewards_gsm8k_1} and Fig.~\ref{fig:rewards_math_1}, when presented with an incorrect reasoning step, SCDPO more accurately identifies the incorrect tokens compared to DPO. Fig.~\ref{fig:rewards_gsm8k_1} shows part of a solution for a GSM8K question. In step 2, the solution incorrectly interprets ``4 less than a dozen'' as ``$4 \times (12 - 4)$'', when it should have been ``$(12 - 4)$''. The SCDPO model correctly highlights ``$4 \times (12 - 4)$'', while the DPO does not. Fig.~\ref{fig:rewards_math_1} shows part of a solution for a MATH question. The solution sums the terms in the expression when two of the terms should have been multiplied. SCDPO correctly highlights the incorrect solution, while DPO does not. These examples show that the stepwise supervision provided in SCDPO results in a better token-level understanding of reasoning errors.

\section{Experiments}

In this section, we first perform a comprehensive empirical comparison between SCDPO and DPO on three kinds of Mistral-7B SFT models. Then, we increase the data used in SFT, DPO, and SCDPO training, using InternLM2-20B as the foundation model, demonstrating the great potential of our method.

\subsection{Comparison using 7B Models}

\begin{table}[t]\fontsize{10}{9}\selectfont
\centering
\caption{Effect of using Step-Controlled DPO (SCDPO) on three different SFT models: a Mistral-7B model finetuned with code-integrated solutions we collected from GPT-4 Code Interpreter, the MetaMath-Mistral-7B model, and MathCoder-Mistral-7B model we finetuned using the MathCoderInstruct dataset, compared to DPO. ``(data-equal)'' denote the DPO baseline using the same amount of data as SCDPO. ``GS'' and ``MA'' are short for GSM8K and MATH respectively.}
\begin{tabularx}{\columnwidth}{>{\raggedright\arraybackslash}X *{2}{>{\centering\arraybackslash\hsize=.6\hsize}X}*{5}{>{\centering\arraybackslash\hsize=.35\hsize}X} }
\toprule
\multirow{2}{*}{\textbf{Method}}   & \multirow{2}{*}{\textbf{GSM8K}} &   \multirow{2}{*}{\textbf{MATH}} & \multicolumn{5}{c}{\textbf{Data}}  \\
\cmidrule(r){4-8} 
&  &  &  \textbf{GS$_\text{dpo}$}& \textbf{MA$_\text{dpo}$}& \textbf{GS$_\text{scdpo}$} &\textbf{MA$_\text{scdpo}$} & \textbf{SFT}  \\
\midrule
\midrule
\multicolumn{8}{c}{Mistral-7B-Ours} \\
\midrule
SFT  & 76.8\%  &  43.2\%     & 	-	& - & - & - & 81K  \\
DPO  & 78.8\%  &  45.1\%   & 	7K	& 5K & - & - &  - \\
DPO\tiny(data-equal)  & 79.0\%  &  45.7\%  & 	13K	& 17K & - & - & -  \\
SCDPO  & \textbf{80.1\%}  &  \textbf{47.7\%}& 	7K	& 5K & 6K & 12K &  - \\
\midrule
\midrule
\multicolumn{8}{c}{MetaMath-Mistral-7B} \\
\midrule
SFT  & 77.7\%  &  28.2\% & 	-	& - & - & - & 395K   \\
DPO  & 81.0\%  &  28.7\%    & 	7K	& 6K & - & - &  -  \\
DPO\tiny(data-equal)  & 81.4\%  &  29.0\%   & 	13K	& 17K & - & - & -  \\
SCDPO  & \textbf{81.7\%}  &  \textbf{29.3\%} & 	7K	& 6K & 6K & 11K &  - \\
\midrule
\midrule
\multicolumn{8}{c}{MathCoder-Mistral-7B} \\
\midrule
SFT  & 78.1\%  &  39.3\%  & 	-	& - & - & - & 80K  \\
DPO  & 79.2\%  &  42.9\%  & 	6K	& 6K & - & - & -  \\
DPO\tiny(data-equal)  & 78.3\%  &  41.1\%  & 12K	& 19K & - & - & - \\
SCDPO  & \textbf{80.4\%}  &  \textbf{43.3\%} & 	6K	& 6K & 6K & 13K & - \\
\bottomrule
\end{tabularx}

\label{tab:sft_dpo_scdpo_performance}
\end{table}

\begin{figure*}[t]
    \centering
    \includegraphics[width=1.02\textwidth]{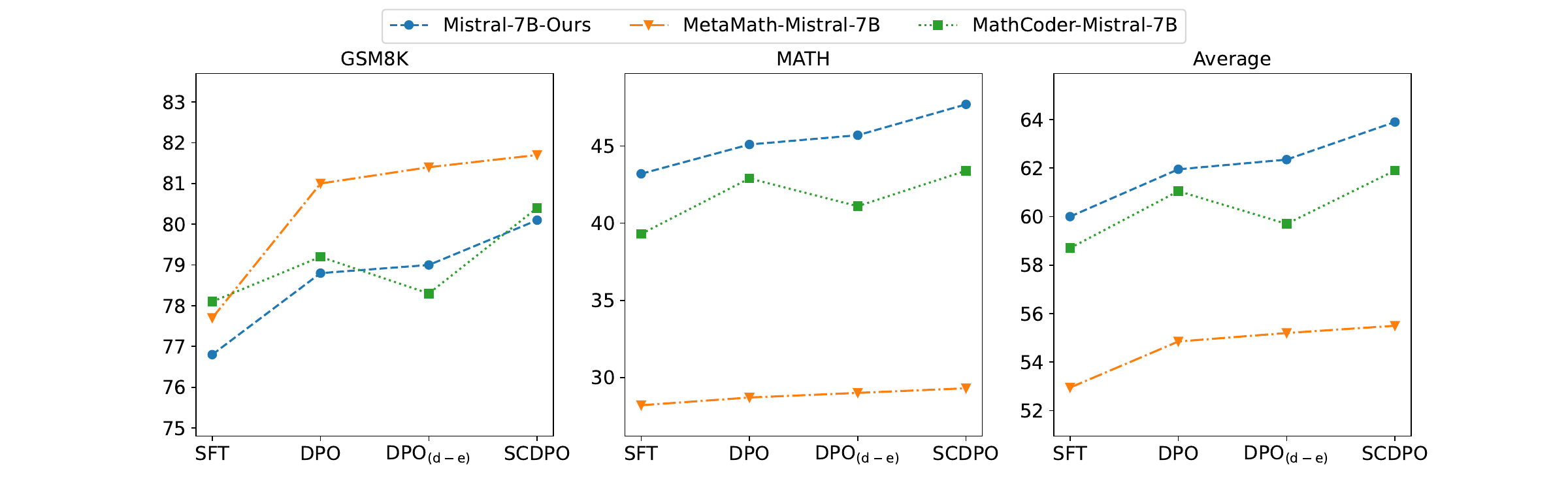}
    \caption{Comparison between SCDPO and DPO. On all three models, the SCDPO method achieves the best performance. Here ``(d-e)'' means data-equal, denoting DPO using the same amount of data as SCDPO.}
    
\label{fig:sft_dpo_scdpo_performance}
\end{figure*}

\textbf{Baseline Models.} We introduce three baseline SFT models: Mistral-7B-Ours, MetaMath-Mistral-7B, and MathCoder-Mistral-7B. All three SFT models use Mistral-7B as the foundation model. Mistral-7B-Ours is finetuned with a math problem-solution dataset we created by collecting multiple solutions from the GPT-4 Code Interpreter for each problem in the GSM8K and MATH training sets and retaining those reaching the correct final answer. This SFT dataset contains 34K question-solution pairs from GSM8K, and 47K from MATH. MetaMath-Mistral-7B is downloaded from the MetaMath HuggingFace repository\footnote{\url{https://huggingface.co/meta-math/MetaMath-Mistral-7B}}. The model is reported to have been trained on the 395K MetaMathQA dataset~\cite{yu2023metamath}, and we do not do any further SFT training on it. MathCoder-Mistral-7B is finetuned using the MathCodeInstruct dataset~\cite{wang2023mathcoder}, downloaded from HuggingFace\footnote{\url{https://huggingface.co/datasets/MathLLMs/MathCodeInstruct}}.

\textbf{Implementation Details.} The supervised finetuning of Mistral-7B-Ours and MathCoder-Mistral-7B is conducted with a learning rate of $1.0\times10^{-5}$ for 3 epochs, with a context length of 2048 tokens. DPO and SCDPO of Mistral-7B-Ours and MathCoder-Mistral-7B are trained with a learning rate of $1.0\times10^{-7}$ for 2 epochs, with a context length of 1024 tokens and $\beta$ set as 0.1. DPO and SCDPO of MetaMath-Mistral-7B are trained with a learning rate of $1.0\times10^{-7}$ for 2 epochs, with a context length of 2048 tokens and $\beta$ set as 0.5. The details of data composition are shown in Tab.~\ref{tab:sft_dpo_scdpo_performance}. The training code is implemented based on HuggingFace's alignment-handbook repository\footnote{\url{https://github.com/huggingface/alignment-handbook}}. The models are trained on 8 NVIDIA A800 80GB GPUs with a batch size of 64.

\textbf{Comparison between SCDPO and DPO.} The results of SFT, DPO, and SCDPO on GSM8K and MATH are shown in Tab.~\ref{tab:sft_dpo_scdpo_performance}. We perform two DPO experiments, with different amounts of training data. One is trained with the $D_\text{naive}$ part (as explained in Sec.~\ref{sec:scdpo_data_collection}) of the SCDPO training data. The other, DPO$_\text{(data-equal)}$, is trained on data expanded from $D_\text{naive}$ to include more preferred-dispreferred DPO training pairs, resulting in a training dataset consisting of approximately the same amount of samples as SCDPO's training dataset. This is to rule out the possibility that the performance gain of SCDPO is the effect of more training samples. As demonstrated in Tab~\ref{tab:sft_dpo_scdpo_performance} and Fig.~\ref{fig:sft_dpo_scdpo_performance}, on all three SFT baseline models, SCDPO shows superior performance compared to DPO. This can be attributed to SCDPO's more detailed supervision on the reasoning steps of the math solutions, demonstrating the effectiveness of our method.

\subsection{Scaling of Data Amount on 20B Model}

\begin{table}[t]\fontsize{9}{11}\selectfont
\centering
\caption{Performance of open-source and closed-source models on two English datasets, GSM8K and MATH, and three Chinese datasets, APE210K, CMATH, and MGSM-zh. All results reported are based on greedy decoding. The best models are marked in \textbf{bold}, and the second best models are \underline{underlined}.}
\begin{tabularx}{\columnwidth}{>{\raggedright\arraybackslash\hsize=2.5\hsize}X >{\centering\arraybackslash\hsize=.3\hsize}X *{4}{>{\centering\arraybackslash\hsize=.6\hsize}X} >{\centering\arraybackslash\hsize=\hsize}X}
\toprule
\multirow{2}{*}{\textbf{Model}} & \multirow{2}{*}{\textbf{Size}} & \multicolumn{2}{c}{\textbf{English}}  & \multicolumn{3}{c}{\textbf{Chinese}} \\
\cmidrule(r){3-4} \cmidrule(r){5-7} 
 &  & \textbf{GSM8K} &   \textbf{MATH}  & \textbf{APE210K} &  \textbf{CMATH} & \textbf{MGSM-zh} \\
\midrule
\midrule
\multicolumn{7}{c}{Closed-Source Models} \\
\midrule
GPT-3.5 & - & 80.8\% & 34.1\% & - & 73.8\% & - \\
GPT-4~\cite{OpenAI2023GPT4TR} & - & \underline{93.6\%} &  \underline{53.6\%} & \underline{84.2\%} &  \textbf{89.3\%} & - \\
GPT-4 Code Interpreter~\cite{zhou2023solving} & - & \textbf{97.0\%} & \textbf{69.7\%} & - & - & - \\
GLM-4~\footnote{\url{https://open.bigmodel.cn/dev/api\#glm-4}} & - & 91.8\% & 49.0\% & \textbf{93.5\%} & \underline{89.0\%} & - \\
Baichuan-3 & - & 88.2\% & 49.2\% & - & - & - \\
\midrule
\midrule
\multicolumn{7}{c}{Open-Source Models} \\
\midrule
Math-Shepherd~\cite{wang2023math} & 7B & 84.1\% & 33.0\% & - & - & - \\
SeaLLM-v2~\cite{seallm} & 7B & 78.2\% & 27.5\% & - & - & 64.8\% \\
DeepSeekMath-RL~\cite{shao2024deepseekmath} & 7B & 86.7\% & \textbf{58.8\%} & 71.9\% & 87.6\%  & \underline{78.4\%} \\
Skywork-13B-Math~\cite{yang2023skymath} & 13B & 72.3\% & 17.0\% & 74.4\% & 77.3\% & - \\
InternLM2-Math~\cite{ying2024internlm} & 20B & 80.7\% & 54.3\% & - & - & - \\
MathGenie~\cite{lu2024mathgenie} & 20B & \underline{87.7\%} & 55.7\% & - & - & - \\
ChatGLM3-32B-RFT-DPO~\cite{xu2024chatglm} & 32B & 82.6\% & 40.6\% & \textbf{89.4\%} & 85.6\% & - \\
Yi-Chat~\cite{Yi} & 34B & 76.0\% & 15.9\% & 65.1\% & 77.7\% & - \\
ToRA~\cite{gou2023tora} & 34B & 80.7\% & 50.8\% & - & 53.4\% & 41.2\% \\
MAmmoTH~\cite{yue2023mammoth} & 70B & 76.9\% & 41.8\% & - & - & - \\
MathCoder~\cite{wang2023mathcoder} & 70B & 83.9\% & 45.1\% & - & - & - \\
WizardMath-v1.0~\cite{luo2023wizardmath} & 70B & 81.6\% & 22.7\% & - & 65.4\% & 64.8\% \\
Qwen~\cite{bai2023qwen} & 72B & 78.9\% & 35.2\% & 77.1\% & 88.1\% & - \\
\midrule
\textbf{InternLM2-SFT} & 20B & 86.4\% & 55.8\% & 77.1\% & 88.4\% & 74.8\% \\
\textbf{InternLM2-SFT-DPO} & 20B & 87.0\% & 57.6\% & 78.7\% & \underline{89.9\%} & 76.0\% \\
\textbf{InternLM2-SFT-DPO$_\text{(data equal)}$} & 20B & 88.2\% & 57.5\% & 78.8\% & 89.3\% & 76.0\% \\
\textbf{InternLM2-SFT-SCDPO} & 20B & \textbf{88.5\%} & \underline{58.1\%} & \underline{79.3\%} & \textbf{90.3\%} & \textbf{80.4\%} \\
\bottomrule
\end{tabularx}

\label{tab:main_results}
\end{table}

\textbf{Training Data.} We increase the amount of SFT training data by collecting solutions for questions in the training set of APE210K~\cite{zhao2020ape210k} from GPT-4 Code Interpreter. APE210K is a dataset containing high-quality Chinese math word problems. After removing the solutions that reach incorrect final answers, we get 169K question-solution pairs. Combining the newly collected data with the original 34K GSM8K data and 47K MATH data, we get an SFT dataset of 250K question-solution pairs. The SCDPO and DPO training data is collected as described before in Sec.~\ref{sec:scdpo_data_collection}.
% When performing SCDPO, we collect 7K GSM8K, 23K MATH, and 21K APE naive DPO preferred-dispreferred pairs in $D_\text{naive}$, and 6K GSM8K, 24K MATH, and 8K APE SCDPO preferred-dispreferred pairs in $D_\text{SC}$. 

\textbf{Training Settings.} We use InternLM2-20B~\cite{cai2024internlm2} as the foundation model, as it has demonstrated high performance in previous works~\cite{lu2024mathgenie,cai2024internlm2}, even surpassing larger models such as Mixtral-8x7B~\cite{jiang2024mixtral} and Llama2-70B~\cite{touvron2023llama} in some cases. In the SFT stage, we finetune the model with a learning rate of $1.0\times10^{-5}$ for 3 epochs, with a context length of 2048 tokens. In DPO and SCDPO training, we use a learning rate of $1.5\times10^{-7}$ to train the SFT model for 2 epochs, with a context length of 1024 and $\beta$ set to 0.1. The models are trained on 16 NVIDIA A800 80GB GPUs with a batch size of 64.

\textbf{Evaluation Datasets.} Five representative mathematical datasets are used in evaluating the models: GSM8K~\cite{cobbe2021training}, MATH~\cite{hendrycks2021measuring}, APE210K~\cite{zhao2020ape210k}, CMATH~\cite{wei2023cmath}, and MGSM-zh~\cite{mgsm}. GSM8K and MATH consist of English math questions, while APE210K, CMATH, and MGSM-zh are consisted of Chinese math questions. The evaluation datasets contain a wide range of problem types, covering mathematical problems from grade-school level to college level, comprehensively evaluating the models' mathematical reasoning abilities. We use greedy decoding for all evaluations.

\textbf{Baselines.} We compare our 20B models with powerful closed-source models such as GPT-3.5~\citep{brown2020language}, GPT-4~\cite{OpenAI2023GPT4TR}, GPT-4 Code Interpreter~\cite{OpenAI2023GPT4TR}, GLM-4~\footnote{\url{https://open.bigmodel.cn/dev/api\#glm-4}}, and Baichuan-3~\footnote{\url{https://www.baichuan-ai.com}}, as well as open-source models such as DeepSeekMath-RL~\cite{shao2024deepseekmath}, Math-Shepherd~\cite{wang2023math}, SeaLLM-v2~\cite{seallm}, Skywork-13B-Math~\cite{yang2023skymath}, InternLM2-Math~\footnote{\url{https://github.com/InternLM/InternLM-Math}}~\cite{ying2024internlm}, MathGenie~\cite{lu2024mathgenie}, ChatGLM3-32B-RFT-DPO~\cite{xu2024chatglm}, Yi-Chat~\cite{Yi}, ToRA~\cite{gou2023tora}, MAmmoTH~\cite{yue2023mammoth}, MathCoer~\cite{wang2023mathcoder}, WizardMath~\cite{luo2023wizardmath}, and Qwen~\cite{bai2023qwen}.

\textbf{Main Results.} Tab.~\ref{tab:main_results} displays our main results, as well as various closed-source and open-source baselines. Our model achieves a score of 88.5\% on GSM8K, 90.3\% on CMATH, and 80.4\% on MGSM-zh, surpassing all
models with published parameters, and obtaining second-best scores among open-source models on MATH and APE210K, with a score of 58.1\% on MATH and 79.3\% on APE210K. While our model rivals the performance of GPT-3.5 and Baichuan-3 on GSM8K and MATH, and surpasses GPT-4 and GLM-4 on MATH, it still underperforms GPT-4 Code Interpreter on GSM8K and MATH, and GLM-4 on APE210K.

Compared to InternLM2-SFT, InternLM2-SFT-SCDPO consistently increases the score on each of the five datasets by approximately 2\% to 3\%. Compared to both InternLM2-SFT-DPO, which uses the $D_\text{naive}$ part of InternLM2-SFT-SCDPO's training data, and InternLM2-SFT-DPO$_\text{(data-equal)}$, which uses about the same amount of training data as InternLM2-SFT-SCDPO, InternLM2-SFT-SCDPO consistently achieves the best performance across all five datasets, highlighting the effectiveness of SCDPO in enhancing mathematical problem-solving abilities.

\section{Limitations and Future Work}

Our work contains the following limitations, and we leave them for future work. Firstly, our work is conducted on purely linguistic models, which struggle to solve mathematical problems requiring an understanding of images. For example, questions in the geometry subject of the MATH dataset exhibit lower accuracy compared to questions in other subjects. A possible solution would be to utilize multimodal techniques, to produce models that can be evaluated with multimodal reasoning datasets~\cite{lu2023mathvista, wang2024measuring}. Secondly, due to the stepwise attribute of SCDPO, it is not very effective on solution formats consisting of pure code. It only works on solutions consisting of natural language chain of thought or interleaved natural language and code. A method to properly enhance pure code solutions needs to be derived, which we leave for future work. Thirdly, as with all language models, our models can potentially generate hallucinations or produce misleading solutions, which can have a negative effect. This can be mitigated with methods such as verification, which we also leave for future work.

\section{Conclusion}

In this work, we propose Step-Controlled DPO (SCDPO), a method to automatically introduce stepwise error supervision to the process of DPO training by generating dispreferred samples that start making errors at a specified step. SCDPO effectively enhances the mathematical reasoning abilities of LLMs. We conduct experiments on three different 7B SFT models, consistently improving the models' performance on mathematical problem-solving tasks and demonstrating the effectiveness and robustness of our method. The 20B model trained with SCDPO on both English and Chinese data achieves the highest score among open-source models on GSM8K, CMATH and MGSM-zh, and second-best score on MATH and APE210K, demonstrating the significant potential of our method.

\newpage
\bibliographystyle{abbrv}
\bibliography{ref}

\newpage
\appendix

% \hx{The appendix should be written at the front of the checklist}

\section{Pilot Study Regarding Correlation Between Temperature and Accuracy}
\label{sec:different_temp_acc}

\begin{figure*}[t]
    \centering
    \includegraphics[width=\linewidth]{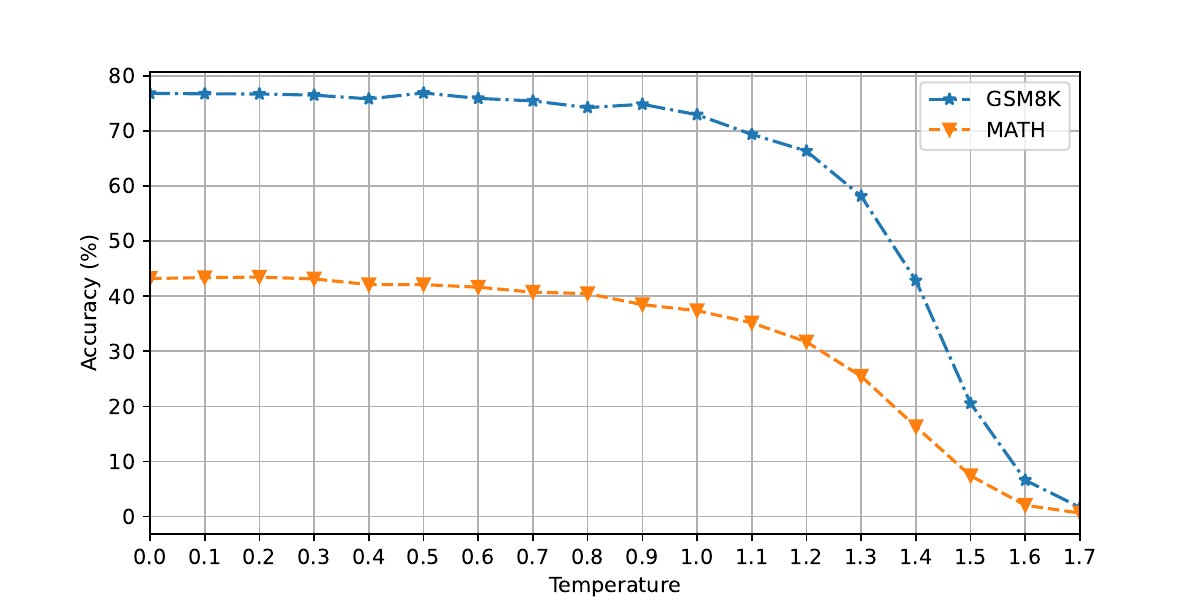}
    \caption{Accuracy of Mistral-7B-Ours (SFT) on GSM8K and MATH when temperature is set at different values.}
    
\label{fig:different_temp_acc}
\end{figure*}

In this section, we provide the results of the SFT version of Mistral-7B-Ours on GSM8K and MATH when temperature is set to different values (from 0.0 to 1.7) as a pilot study. As demonstrated in Fig.~\ref{fig:different_temp_acc}, with the increase of temperature, accuracy shows a trend of decreasing. When the temperature is between 0.0 and 1.0, the accuracy is relatively stable. When the temperature is higher than 1.0, accuracy on the two datasets starts to degrade, as errors are more likely to occur. This rise in the occurrence of errors can be used to create erroneous reasoning steps in SCDPO.

\section{Further Credit Assignment Analysis Examples}
\label{sec:credit_assignment_examples}

\begin{figure*}[t]
    \centering
    \includegraphics[width=\linewidth]{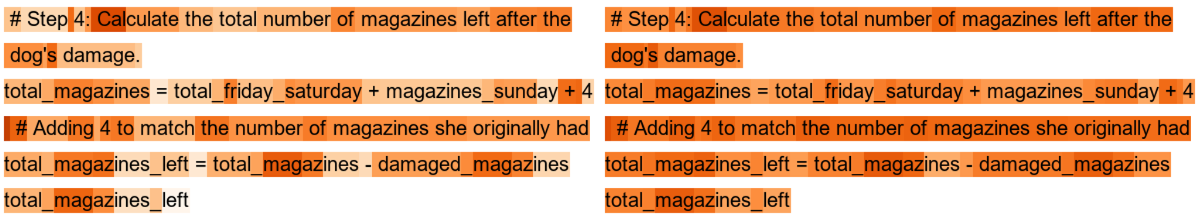}
    \caption{Credit assignment of part of a solution for a GSM8K problem. Each token is colored corresponding to the DPO implicit reward as expressed in Eq.~\ref{eq:reward_param} (darker is higher). The left is the credit assignment of SCDPO, which correctly highlighted the error -- the number of damaged magazines (which is 4) should not be first added to and then extracted from ``total\_magazines'', while the credit assignment of DPO on the right fails to highlight it.}
    
\label{fig:rewards_gsm8k_2}
\end{figure*}

\begin{figure*}[t]
    \centering
    \includegraphics[width=\linewidth]{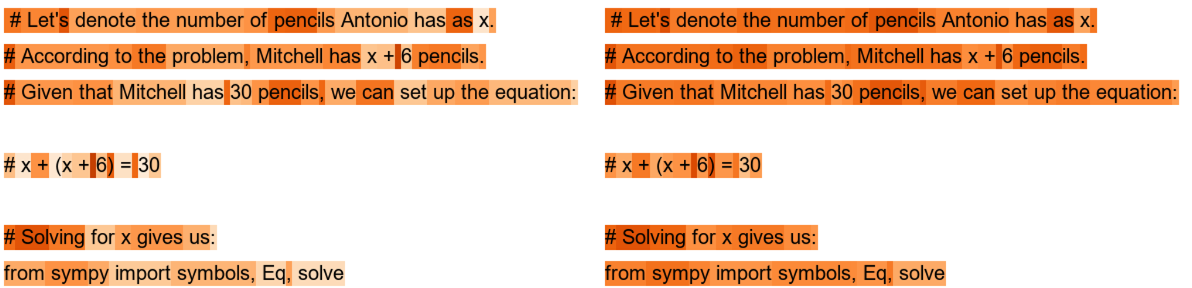}
    \caption{Credit assignment of part of a solution for a GSM8K problem. Each token is colored corresponding to the DPO implicit reward as expressed in Eq.~\ref{eq:reward_param} (darker is higher). The left is the credit assignment of SCDPO, which correctly highlighted the error -- Mitchell has 30 pencils, and Antonio has 6 less pencils than Michell, which is $30 - 6$, so the introduction of $x$ is not needed, and $x + (x 
 + 6) = 30$ is incorrect, while the credit assignment of DPO on the  right fails to highlight it.}
    
\label{fig:rewards_gsm8k_3}
\end{figure*}

\begin{figure*}[t]
    \centering
    \includegraphics[width=\linewidth]{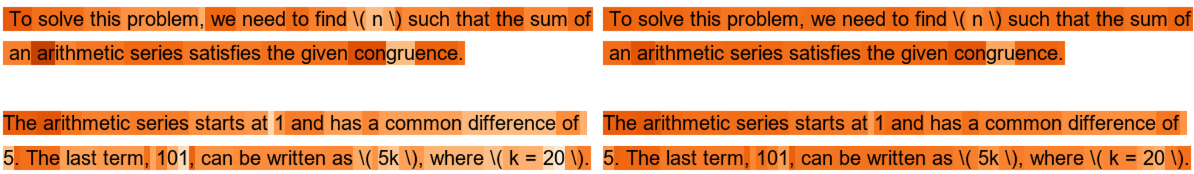}
    \caption{Credit assignment of part of a solution for a MATH problem. Each token is colored corresponding to the DPO implicit reward as expressed in Eq.~\ref{eq:reward_param} (darker is higher). The left is the credit assignment of SCDPO, which correctly highlighted the error -- $101$ cannot be written as $5k$ where $k=20$, while the credit assignment of DPO on the  right fails to highlight the error.}
    
\label{fig:rewards_math_2}
\end{figure*}

In this section, we present several credit assignment analysis examples, comparing SCDPO to DPO. Fig.~\ref{fig:rewards_gsm8k_2}, Fig.~\ref{fig:rewards_gsm8k_3} and Fig.~\ref{fig:rewards_math_2} show examples of part of the solutions of questions taken from GSM8K and MATH datasets, colored with the DPO implicit reward of each token (darker is higher). As demonstrated in the examples, SCDPO is better than DPO at identifying the errors in the reasoning steps.

\newpage
\section*{NeurIPS Paper Checklist}

%%% BEGIN INSTRUCTIONS %%%
The checklist is designed to encourage best practices for responsible machine learning research, addressing issues of reproducibility, transparency, research ethics, and societal impact. Do not remove the checklist: {\bf The papers not including the checklist will be desk rejected.} The checklist should follow the references and follow the (optional) supplemental material.  The checklist does NOT count towards the page
limit. 

Please read the checklist guidelines carefully for information on how to answer these questions. For each question in the checklist:
\begin{itemize}
    \item You should answer \answerYes{}, \answerNo{}, or \answerNA{}.
    \item \answerNA{} means either that the question is Not Applicable for that particular paper or the relevant information is Not Available.
    \item Please provide a short (1–2 sentence) justification right after your answer (even for NA). 
   % \item {\bf The papers not including the checklist will be desk rejected.}
\end{itemize}

{\bf The checklist answers are an integral part of your paper submission.} They are visible to the reviewers, area chairs, senior area chairs, and ethics reviewers. You will be asked to also include it (after eventual revisions) with the final version of your paper, and its final version will be published with the paper.

The reviewers of your paper will be asked to use the checklist as one of the factors in their evaluation. While "\answerYes{}" is generally preferable to "\answerNo{}", it is perfectly acceptable to answer "\answerNo{}" provided a proper justification is given (e.g., "error bars are not reported because it would be too computationally expensive" or "we were unable to find the license for the dataset we used"). In general, answering "\answerNo{}" or "\answerNA{}" is not grounds for rejection. While the questions are phrased in a binary way, we acknowledge that the true answer is often more nuanced, so please just use your best judgment and write a justification to elaborate. All supporting evidence can appear either in the main paper or the supplemental material, provided in appendix. If you answer \answerYes{} to a question, in the justification please point to the section(s) where related material for the question can be found.

IMPORTANT, please:
\begin{itemize}
    \item {\bf Delete this instruction block, but keep the section heading ``NeurIPS paper checklist"},
    \item  {\bf Keep the checklist subsection headings, questions/answers and guidelines below.}
    \item {\bf Do not modify the questions and only use the provided macros for your answers}.
\end{itemize}

%%% END INSTRUCTIONS %%%

\begin{enumerate}

\item {\bf Claims}
    \item[] Question: Do the main claims made in the abstract and introduction accurately reflect the paper's contributions and scope?
    \item[] Answer: \answerYes{} % Replace by \answerYes{}, \answerNo{}, or \answerNA{}.
    \item[] Justification: The main claims made in the abstract and introduction accurately reflect the paper's contributions and scope.

\item {\bf Limitations}
    \item[] Question: Does the paper discuss the limitations of the work performed by the authors?
    \item[] Answer: \answerYes{} % Replace by \answerYes{}, \answerNo{}, or \answerNA{}.
    \item[] Justification: The limitations of the work is discussed in the Limitations and Future Work section of the paper.

\item {\bf Theory Assumptions and Proofs}
    \item[] Question: For each theoretical result, does the paper provide the full set of assumptions and a complete (and correct) proof?
    \item[] Answer: \answerYes{} % Replace by \answerYes{}, \answerNo{}, or \answerNA{}.
    \item[] Justification: For each theoretical result, the paper provide the full set of assumptions and a complete proof.

    \item {\bf Experimental Result Reproducibility}
    \item[] Question: Does the paper fully disclose all the information needed to reproduce the main experimental results of the paper to the extent that it affects the main claims and/or conclusions of the paper (regardless of whether the code and data are provided or not)?
    \item[] Answer: \answerYes{} % Replace by \answerYes{}, \answerNo{}, or \answerNA{}.
    \item[] Justification: We disclose all the information needed to reproduce the main experimenta results of the paper in the Experiments section.

\item {\bf Open access to data and code}
    \item[] Question: Does the paper provide open access to the data and code, with sufficient instructions to faithfully reproduce the main experimental results, as described in supplemental material?
    \item[] Answer: \answerNo{} % Replace by \answerYes{}, \answerNo{}, or \answerNA{}.
    \item[] Justification: We plan to provide open access to the data and code upon acceptance of the paper.

\item {\bf Experimental Setting/Details}
    \item[] Question: Does the paper specify all the training and test details (e.g., data splits, hyperparameters, how they were chosen, type of optimizer, etc.) necessary to understand the results?
    \item[] Answer: \answerYes{} % Replace by \answerYes{}, \answerNo{}, or \answerNA{}.
    \item[] Justification: We specify all the training and test details in the Experiments section.

\item {\bf Experiment Statistical Significance}
    \item[] Question: Does the paper report error bars suitably and correctly defined or other appropriate information about the statistical significance of the experiments?
    \item[] Answer: \answerNo{} % Replace by \answerYes{}, \answerNo{}, or \answerNA{}.
    \item[] Justification: Error bars are not reported because it would be too computationally expensive for LLM training.

\item {\bf Experiments Compute Resources}
    \item[] Question: For each experiment, does the paper provide sufficient information on the computer resources (type of compute workers, memory, time of execution) needed to reproduce the experiments?
    \item[] Answer: \answerYes{} % Replace by \answerYes{}, \answerNo{}, or \answerNA{}.
    \item[] Justification: The computer resources needed to reproduce the experiments are explained in the Experiments section.
    
\item {\bf Code Of Ethics}
    \item[] Question: Does the research conducted in the paper conform, in every respect, with the NeurIPS Code of Ethics \url{https://neurips.cc/public/EthicsGuidelines}?
    \item[] Answer: \answerYes{} % Replace by \answerYes{}, \answerNo{}, or \answerNA{}.
    \item[] Justification: The research conducted in the paper conform, in every respect, with the NeurIPS Code of Ethics.

\item {\bf Broader Impacts}
    \item[] Question: Does the paper discuss both potential positive societal impacts and negative societal impacts of the work performed?
    \item[] Answer: \answerYes{} % Replace by \answerYes{}, \answerNo{}, or \answerNA{}.
    \item[] Justification: Potential positive societal impacts and negative societal impacts of the work performed are discussed in the Limitations and Future Work section.
    
\item {\bf Safeguards}
    \item[] Question: Does the paper describe safeguards that have been put in place for responsible release of data or models that have a high risk for misuse (e.g., pretrained language models, image generators, or scraped datasets)?
    \item[] Answer: \answerNA{} % Replace by \answerYes{}, \answerNo{}, or \answerNA{}.
    \item[] Justification: Our work is limited to mathematical problem solving and contains no pretraining or scraped datasets. So the risk for misuse is small.

\item {\bf Licenses for existing assets}
    \item[] Question: Are the creators or original owners of assets (e.g., code, data, models), used in the paper, properly credited and are the license and terms of use explicitly mentioned and properly respected?
    \item[] Answer: \answerYes{} % Replace by \answerYes{}, \answerNo{}, or \answerNA{}.
    \item[] Justification: The creators or original owners of assets (e.g., code, data, models), used in the paper, are properly credited and the license and terms of use are explicitly mentioned and properly respected.

\item {\bf New Assets}
    \item[] Question: Are new assets introduced in the paper well documented and is the documentation provided alongside the assets?
    \item[] Answer: \answerNA{} % Replace by \answerYes{}, \answerNo{}, or \answerNA{}.
    \item[] Justification: The new assets introduced in the paper are not released upon submission, and will only be released upon the acceptance of the paper. 

\item {\bf Crowdsourcing and Research with Human Subjects}
    \item[] Question: For crowdsourcing experiments and research with human subjects, does the paper include the full text of instructions given to participants and screenshots, if applicable, as well as details about compensation (if any)? 
    \item[] Answer: \answerNA{} % Replace by \answerYes{}, \answerNo{}, or \answerNA{}.
    \item[] Justification: The paper does not involve crowdsourcing nor research with human subjects.

\item {\bf Institutional Review Board (IRB) Approvals or Equivalent for Research with Human Subjects}
    \item[] Question: Does the paper describe potential risks incurred by study participants, whether such risks were disclosed to the subjects, and whether Institutional Review Board (IRB) approvals (or an equivalent approval/review based on the requirements of your country or institution) were obtained?
    \item[] Answer: \answerNA{} % Replace by \answerYes{}, \answerNo{}, or \answerNA{}.
    \item[] Justification: The paper does not involve crowdsourcing nor research with human subjects.

\end{enumerate}

\end{document}